\documentclass[11pt]{article}
\usepackage[hyperref]{ccl2025-en}
\usepackage{times}
\usepackage{url}
\usepackage{latexsym}
\usepackage{fancyhdr}
\usepackage{amsmath}
\usepackage{threeparttable}

\usepackage{algorithm}
\usepackage{algorithmicx}
\usepackage{algpseudocode}
\usepackage{amsmath}

\usepackage{graphicx}
\title{System Report for CCL25-Eval Task 10: \protect\\ Prompt-Driven Large Language Model Merge for Fine-Grained Chinese Hate Speech Detection}

\author{
    Binglin Wu\textsuperscript{\dag}, 
    Jiaxiu Zou\textsuperscript{\dag}, 
    Xianneng Li\textsuperscript{*} \\
    School of Economics and Management, Dalian University of Technology \\
    xianneng@dlut.edu.cn\\
    }

\date{}
\begin{document}
\maketitle
\begin{abstract}
  The proliferation of hate speech on Chinese social media poses urgent societal risks, yet traditional systems struggle to decode context-dependent rhetorical strategies and evolving slang. To bridge this gap, we propose a novel three-stage LLM-based framework: \textbf{Prompt Engineering}, \textbf{Supervised Fine-tuning}, and \textbf{LLM Merging}. First, context-aware prompts are designed to guide LLMs in extracting implicit hate patterns. Next, task-specific features are integrated during supervised fine-tuning to enhance domain adaptation. Finally, merging fine-tuned LLMs improves robustness against out-of-distribution cases. Evaluations on the STATE-ToxiCN benchmark validate the framework's effectiveness, demonstrating superior performance over baseline methods in detecting fine-grained hate speech.
\end{abstract}

\cclfootnote{
    %
    %
    \hspace{-0.65cm}  
    \dag Equal Contribution\\
    *Corresponding Author\\
    \textcopyright 2025 China National Conference on Computational Linguistics
    
    \noindent Published under Creative Commons Attribution 4.0 International License\\

}

\section{Introduction}
\label{intro}

The rapid growth of social media platforms has led to a global surge in online hate speech, which not only inflicts psychological harm on targeted individuals or groups but also exacerbates social tensions and fuels collective antagonism \cite{arora2023detecting}. While existing technologies can preliminarily detect explicit hate content \cite{schmidt2017survey}, Chinese hate expressions are often characterized by implicitness, diversity, and context-dependency \cite{qian-etal-2018-leveraging}. Offensive content may be embedded through metaphors, sarcasm, or indirect references \cite{fortuna2018survey}, frequently targeting specific group attributes such as geography, gender, or ethnicity \cite{mathew2021hatexplain}. Against this backdrop, \textbf{fine-grained hate speech detection} has emerged as a critical research direction to address this issue. It aims to precisely dissect hate elements—such as target entities, arguments, victimized groups, and hate attributes \cite{vidgen-etal-2021-learning}—from textual data, enabling more accurate identification and regulation of online hate speech.

The core requirement of fine-grained hate speech detection lies in models that can not only recognize explicit offensive lexicons but also infer discriminatory intent from contextual semantics \cite{elsherief-etal-2021-latent}, while strictly adhering to structured output specifications \cite{pavlopoulos-etal-2020-toxicity}. However, current mainstream models face three critical bottlenecks:(1)\textbf{Semantic Complexity}: Traditional rule-based or shallow machine learning methods, as well as directly applied large language models, struggle to accurately capture the implicit and diverse fine-grained semantic features inherent in Chinese hate speech \cite{talat2016hateful}.(2)\textbf{Incomplete Information Extraction}: General-purpose pre-trained models lack targeted attention to hate speech components, resulting in incomplete extraction of critical information.(3)\textbf{Generalization Limitations}: Single training paradigms are susceptible to data distribution biases, limiting model generalization in complex scenarios and hindering adaptability to dynamic online environments \cite{gururangan-etal-2020-dont}.

To address these challenges, this study proposes a hybrid training framework based on the Qwen2.5-7B-Instruct LLM \cite{qwen2025qwen25technicalreport}, employing a three-stage optimization strategy. First, Prompt Engineering guides the model to focus on hate speech elements (e.g., victimized group classification and metaphor identification rules) while enforcing structured output through task-oriented templates. Second, Supervised Fine-Tuning (SFT) \cite{ouyang2022training} enhances the model’s ability to parse fine-grained semantics using high-quality annotated data, particularly improving discrimination accuracy for implicit hate expressions. Finally, Model Merging \cite{matena2022merging} innovatively integrates multi-stage models via the LLM Merging method, which sparsifies task vectors by pruning extreme parameters, thereby synthesizing complementary features from different training phases to boost robustness. Experimental results demonstrate stable performance scores of 0.3553 and 0.3555 on preliminary and final test sets, respectively, with over 15\% accuracy improvement in hate detection compared to baseline models. The fused model also exhibits exceptional adaptability in complex scenarios such as multi-group attacks and cross-context generalization. This work provides a theoretically innovative and practically valuable technical pathway for Chinese fine-grained hate speech detection, contributing significantly to fostering safer online discourse environments.
%
%

\section{Methodology}

\subsection{Framework Overview}

\begin{figure}[!htbp]
    \centering
    \includegraphics[width=0.9\textwidth]{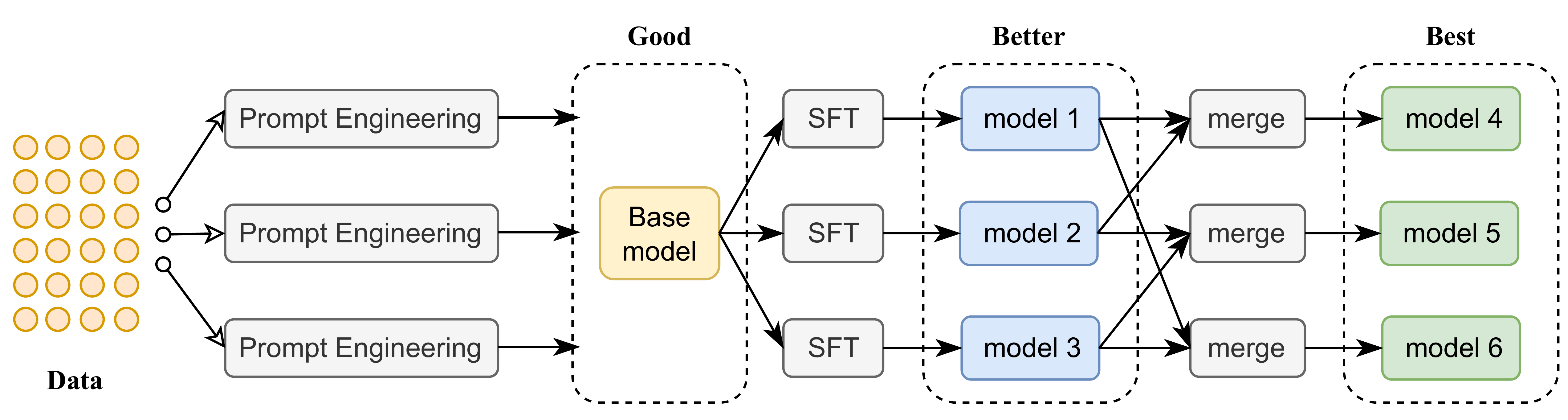}
    \caption{Framework Architecture}
    \label{fig:framework}
\end{figure}

The proposed framework comprises three pivotal components: (1) Domain-specific Prompt Engineering, (2) Task-oriented Supervised Fine-tuning, and (3) Dynamic LLM Merge. As illustrated in the hierarchical architecture of the algorithmic framework figure \ref{fig:framework}, the system operates through phased optimization: prompt engineering guides the model to concentrate on fine-grained hate elements, the supervised fine-tuning phase enhances the model's discriminative capacity for implicit semantic nuances, and model merge enhances both the recognition accuracy and generalization capabilities of the system.

\subsection{Domain-specific Prompt Strategy}

The Prompt Strategy enhances structured output capabilities and fine-grained hate judgment logic through domain-specific prompt template design. Specifically, the prompt template incorporates three core components:

First, it defines clear task objectives by mandating the model to output results following a "four-tuple" structured framework. To reinforce the model’s understanding of this format, contextual examples are strategically embedded immediately after defining each field.Second, it embeds explicit definitions of hate speech while establishing contrasting non-targeted content boundaries through dual-directional examples. For instance, the prompt explicitly contrasts hate speech with non-targeted content, clarifying criteria with phrases like "ordinary information without group targeting does not constitute hate speech." This bidirectional guidance reduces false positives by sharpening the model’s ability to differentiate subtle boundary cases.Third, it optimizes target group extraction by integrating predefined category explanations that map to common social group attributes. The prompt systematically breaks down each category’s defining features, and emphasizes handling overlapping scenarios. This structured approach ensures the model prioritizes contextually relevant group attributes while mitigating oversights in complex expressions.

\begin{figure}[!htbp]
    \centering
    \includegraphics[width=0.9\linewidth]{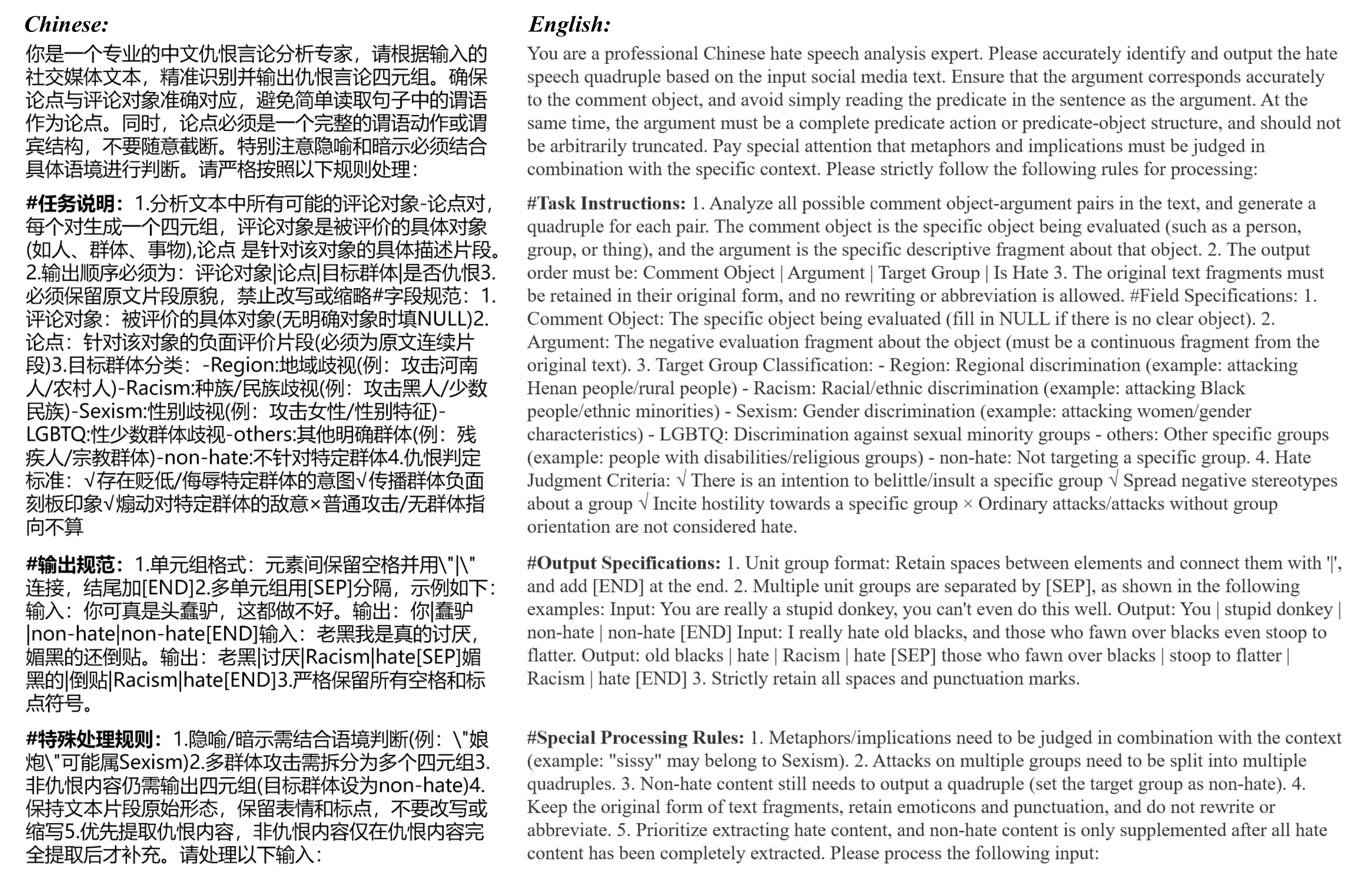}
    \caption{Sample Prompt}
    \label{fig:Sample Prompt}
\end{figure}

Figure \ref{fig:Sample Prompt} is a sample prompt. Through these designs, the prompt shifts the model’s focus from generalized semantic analysis to targeted feature extraction governed by structured guidelines, aligning detection logic with fine-grained hate speech components. The domain-specific prompt not only enforces rigorous output formatting but also embeds implicit reasoning pathways for decoding implicit rhetoric.
\subsection{Task-oriented Supervised Fine-tuning}

Given a pre-trained large language model $\theta_{pre}$ and a labeled dataset $D=\{({{x}_{i}},{{y}_{i}})\}_{i=1}^{N}$ for hate speech detection, full parameter supervised fine-tuning minimizes the loss function:

\begin{equation}
\mathcal{L}(\theta) = -\sum_{i=1}^N \left[y_i \log p(\hat{y_i}|\mathbf{x}i;\theta) + (1-y_i)\log(1-p(\hat{y_i}|\mathbf{x}i;\theta))\right] + \lambda |\theta - \theta_{pre}|^2
\label{eq:loss}
\end{equation}

where $\theta$ denotes the complete set of trainable parameters, $\hat{y_i}$ represents the model's predicted probability for the i-th sample, $\lambda$  serves as the L2 regularization coefficient that governs parameter magnitude constraints to mitigate overfitting. The parameter update rule of the AdamW optimizer is defined as:

\begin{equation}
\theta_{t+1} = \theta_t - \eta \cdot \frac{m_t}{\sqrt{v_t} + \epsilon} - \eta\lambda\theta_t
\label{eq:adamw}
\end{equation}

where $\eta$ denotes the learning rate, $m_t$ and $v_t$  represent the exponentially decaying first and second moment estimates of gradients, respectively, and $\epsilon$ is a small constant ensuring numerical stability.

\subsection{Dynamic Large Language Model Merge}  

The capabilities learned by LLMs fine-tuned with different prompt strategies exhibit significant variations.Recent studies show merging large language models (LLMs) effectively enhances performance and generalization. For instance, in e-commerce intention recognition, merged models demonstrate stronger robustness when processing noisy multimodal data, significantly improving accuracy in complex scenarios \cite{li2025cusmer}.Building upon the methodologies presented in \cite{yadav2023ties,davari2024model}, we propose a LLM Merging algorithm to integrate these diverse capabilities, with the detailed workflow outlined in Algorithm \ref{alg:llm merging}.

Given fine-tuned LLMs $\{\theta_t\}_{t=1}^n$ and a base LLM $\theta_{\text{base}}$, we first construct corresponding task vectors $\tau$. Based on task vectors $\{\tau_t\}_{t=1}^n$, the LLM Merging method proceeds through three sequential steps to achieve parameter merging:
\begin{itemize}
    \item \textbf{Prune:} We partition the model into layers. For each layer, a masking process is implemented to filter out large outliers and minor perturbations, using $\alpha$ and $\beta$ as thresholds for the right-tail (upper bound) and left-tail (lower bound) distributions, respectively. The resulting layer-specific masks ${m}^{\alpha,\beta}_{t,layer}$ are aggregated across all layers to generate the final unified mask ${m}^{\alpha,\beta}_{t}$. The mask is then applied to the task vector $\tau_t$ to derive the refined parameter set $\hat{\tau}_t$, from which we extract the task-specific direction $\hat{\gamma}_t$ and the magnitude of change $\hat{\mu}_t$.
    \item \textbf{Direct:} We construct a directional alignment vector $\gamma_m$ to resolve sign inconsistencies among corresponding parameters across different models.  Specifically, task vectors sharing the same sign direction are aggregated, and the orientation demonstrating the highest cumulative magnitude is selected as the consensus direction.
    \item \textbf{Merge:} For each parameter, we construct the chosen set of task vectors $\mathcal{A}^p$, which only retains the parameter values of the models whose symbolic directions are the same as the consensus direction. Finally, calculate their average values $\tau_m^p$, scale them and then add them to the base parameters to obtain the final merged parameters $\theta_m$.
\end{itemize}

\begin{center}
\begin{minipage}{0.7\linewidth}
\begin{algorithm}[H]
\caption{LLM MERGING Procedure.}
\label{alg:llm merging}
\begin{algorithmic}[1]
\Require Fine-tuned LLMs $\{\theta_t\}_{t=1}^n$, Initialization $\theta_{\text{base}}$, $\alpha$, $\beta$ and $\lambda$.
\Ensure Merged LLM $\theta_m$
\ForAll{$t \in [1, ..., n]$}
    \State $\triangleright$ Create task vectors.
    \State $\tau_t = \theta_t - \theta_{\text{base}}$
    \State $\triangleright$ Step 1: Prune redundant vectors.
    \ForAll{$layer \in Layers(\theta)$}
        \State ${m}^{\alpha}_{t,layer} \gets \text{mask\_top\_k\_percent}(k=\alpha)$
        \State ${m}^{\beta}_{t,layer} \gets \text{mask\_bottom\_k\_percent}(k=\beta)$
        \State ${m}^{\alpha,\beta}_{t,layer} \gets \text{merge\_masks}({m}^{\alpha}_{t},{m}^{\beta}_{t})$
    \EndFor
    \State ${m}^{\alpha,\beta}_{t} \gets \text{stack\_masks}({{\left\{ {{m}^{\alpha,\beta}_{t,layer} } \right\}}_{layer \in Layers}})$
    \State $\hat{\tau}_t \gets {m}^{\alpha,\beta}_{t}\cdot\tau_t$
    \State $\hat{\gamma}_t \gets \text{sgn}(\hat{\tau}_t)$
    \State $\hat{\mu}_t \gets |\hat{\tau}_t|$
\EndFor
\State $\triangleright$ Step 2: Indicate task directions.
\State $\gamma_m = \text{sgn}\left(\sum_{t=1}^n \hat{\gamma}_t\odot\hat{\mu}_t\right)$
\State $\triangleright$ Step 3: Merge chosen task vectors.
\ForAll{$p \in [1, ..., d]$}
    \State $\mathcal{A}^p = \{t \in [n] \mid \hat{\gamma}_t^p = \gamma_m^p\}$
    \State $\tau_m^p \gets \dfrac{1}{|\mathcal{A}^p|} \sum_{t \in \mathcal{A}^p} \hat{\tau}_t^p$
\EndFor
\State Obtain merged checkpoint
\State $\theta_m \gets \theta_{\text{base}} + \lambda * \tau_m$
\State \Return $\theta_m$
\end{algorithmic}
\end{algorithm}
\end{minipage}
\end{center}

\section{Experiments and Results}

\subsection{Dataset}

The STATE-ToxiCN dataset \cite{bai2025state} comprises 8,000 Chinese social media comments (e.g., from Tieba and Zhihu) annotated with fine-grained quadruples (Target | Argument | Targeted Group | Hateful) for hate speech recognition. Each sample captures explicit targets (or NULL), argumentative fragments, affected groups (geographic, race, gender, LGBTQ, other/Non-hate), and binary hate labels, yielding 9,405 quadruples (5,949 hateful, 3,456 non-hate). It supports multi-target annotations via [SEP] separators and enforces full element extraction even for non-hate texts, covering scenarios like racial bias and gender conflicts. Rigorous validation ensures semantic consistency, offering granular supervision for modeling hate speech components beyond sentence-level classification.

\subsection{Evaluation Metrics}
The evaluation metrics consist of the F1-scores for hard matching and soft matching between the submitted results and the standard answers, as well as the average of these two F1-scores. The calculation method is consistent with the scikit-learn library.

For the hard matching, a predicted four-tuple is considered correctly extracted if and only if each element of the predicted four-tuple is completely identical to the corresponding element in the answer.

For the soft matching, a predicted four-tuple is considered correctly extracted under the following conditions: the "Targeted Group" and "Hateful" elements of the predicted four-tuple are completely identical to the corresponding elements in the standard answer, and the string matching degree of the "Target" and "Argument" elements between the predicted four-tuple and the standard answer exceeds 50\% . The similarity is calculated as:
\begin{equation}
    \text{Similarity} = \frac{M \times 2}{\text{len}_{\text{pred}} + \text{len}_{\text{gold}}}
\end{equation}

where $\text{len}_{\text{pred}}$ is the length of the predicted four-tuple, $\text{len}_{\text{gold}}$ is the length of the standard answer, and $M$ is the length of the longest common subsequence between the predicted four-tuple and the standard answer.

The F1-score is calculated as:
\begin{equation}
F1 = 2 \times \frac{P \times R}{P + R}
\end{equation}

where $P$ is precision and $R$ is recall. These metrics comprehensively evaluate the performance of the model from both strict matching and partial matching perspectives.
    
\subsection{Experiment Setup}
The experiments were conducted using 8 NVIDIA RTX 4090 GPUs.  During the training phase, we set the global learning rate to 1e-5, with a total batch size of 64, and trained for 8 epochs. To optimize GPU memory utilization, we employed the DeepSpeed Zero-3 Offload strategy to offload partial model parameters to CPU memory and integrated Flash Attention 2.0 to accelerate attention computation.
\subsection{Experiment Results}
\subsubsection{Overall Comparative Analysis}

The comprehensive evaluation of various post-training approaches on different Qwen2.5 variants is shown on Table \ref{tab:overall}. The RFT method underperformed significantly, likely due to inadequate guidance from its rule-based reward mechanism in capturing nuanced hate speech patterns.  Notably, the CPT+SFT approach applied to the base model demonstrated competitive performance, outperforming direct SFT on the Instruct variant (0.3379 vs. 0.3436). We hypothesize that extended CPT training with additional domain-specific corpora could further enhance this performance gap.

Our proposed method achieves state-of-the-art results across all three metrics on Test1, with particularly notable improvements in Hard Score, indicating superior detection capability for implicit hate expressions. Remarkably, without any task-specific adaptation, our framework maintains robust performance on Test2 (Score: 0.3545), demonstrating both methodological effectiveness and generalization capabilities.

\begin{table}[htbp]
\centering
\caption{Overall results on test1}
\vspace{4pt}
\begin{threeparttable} 
\begin{tabular}{ccc c c}
\hline
Base Model & Method & Score & Hard Score & Soft Score \\
\hline
Qwen2.5-3B-Instruct & RFT & 0.2021 & 0.1126 & 0.2915  \\
Qwen2.5-7B-Base & CPT+SFT & 0.3379 & 0.2353 & 0.4404  \\
Qwen2.5-7B-Instruct & SFT & 0.3436 & 0.2383 & 0.4489  \\
Qwen2.5-7B-Instruct & Ours & \textbf{0.3553} & \textbf{0.2504} & \textbf{0.4604} \\ 
\hline
\end{tabular}
\begin{tablenotes} 
\footnotesize
\item[1] RFT: Reinforcement Fine-tuning based on GRPO Algorithm \cite{shao2024deepseekmath}
\item[2] CPT+SFT: Continue Pre-training using COLD dataset \cite{deng2022cold}. The subsequent SFT uses the prompt strategy of ICL+NH+CE.
\end{tablenotes}
\end{threeparttable}
\label{tab:overall}
\end{table}

\subsubsection{Effect of Prompt Strategy}

The experimental results in Table \ref{tab:prompt} demonstrate a progressive improvement as prompt strategies are incrementally enhanced. The baseline ICL approach achieved a score of 0.2921, while the most comprehensive strategy combining ICL with Non-Hate examples, Category Explanations, and explicit Judge Criteria attained the highest performance. This 17.6\% relative improvement from baseline to the optimal configuration suggests that clarifying detection boundaries through category explanations and judgment criteria significantly enhances model discernment in ambiguous cases. Particularly, the soft score improvement (14.6\% increase) indicates enhanced capability to handle nuanced expressions like sarcasm and homophonic substitutions prevalent in Chinese hate speech.

These results emphasize the importance of combining structured detection guidelines with linguistic and cultural awareness in prompt engineering for Chinese hate speech identification.

\begin{table}[htbp]
\centering
\caption{Results of different prompt strategies on test1}
\vspace{4pt}
\begin{threeparttable} 
\begin{tabular}{ccc c c}
\hline
Prompt Strategy & Score & Hard Score & Soft Score \\
\hline
ICL & 0.2921 & 0.1926 & 0.3916 \\
ICL+Non Hate & 0.3279 & 0.2196 & 0.4362 \\
ICL+NH+Category Explain & 0.3340 & 0.2316 & 0.4365 \\
ICL+NH+CE+Judge Criteria &  \textbf{0.3436} & \textbf{0.2383} & \textbf{0.4489} \\
\hline
\end{tabular}
\begin{tablenotes} 
\footnotesize
\item[1] ICL: In-context Learning, specify the task requirements and provide examples 
\end{tablenotes}
\end{threeparttable}
\label{tab:prompt}
\end{table}

\subsubsection{Effect of LLM Merge}

 Results shown in Table \ref{tab:merge} reveal significant performance enhancements through LLM Merge. Merging base ICL with its enhanced version (ICL+NH) produced Merge1 (0.3412 score), already surpassing the standalone ICL+NH+CE model (0.3340). Subsequent merging iterations demonstrated compounding benefits, with Merge2 (0.3530) and Merge3 (0.3553) progressively outperforming all individual prompt-engineered models, including the comprehensive ICL+NH+CE+JC configuration (0.3436). This 3.4\% improvement from the best single-model to merged models suggests complementary strengths in different detection approaches – where original models might overfit specific patterns, merged versions likely balance categorical understanding from explicit prompts with nuanced judgment capabilities. Notably, the hard score increased 5.1\% (0.2383$\rightarrow$0.2504) through merging, indicating improved consensus on definitive hate speech cases, while the 2.5\% soft score gain (0.4489$\rightarrow$0.4602) reflects enhanced handling of ambiguous expressions. 
 
 However, diminishing returns between Merge2 (0.3530) and Merge3 (0.3553) suggest a potential limit to current merging strategies' effectiveness, possibly requiring novel fusion techniques for Chinese's context-dependent hate markers like dialectal variations and historical allusion.   These results advocate for hybrid approaches combining prompt engineering with model merging to address Chinese hate speech's unique linguistic and cultural complexity.

\begin{table}[htbp]
\centering
\caption{Results of merged models on test1}
\vspace{4pt}
\begin{threeparttable} 
\begin{tabular}{ccc c c}
\hline
Model & Score & Hard Score & Soft Score \\
\hline
ICL & 0.2921 & 0.1926 & 0.3916 \\
ICL+Non Hate & 0.3279 & 0.2196 & 0.4362 \\
\textbf{ICL\&ICL+NH (Merge1)} & \textbf{0.3412} & \textbf{0.2358} & \textbf{0.4467} \\
ICL+NH+Category Explain & 0.3340 & 0.2316 & 0.4365 \\
\textbf{Merge1\&ICL+NH+CE (Merge2)} & \textbf{0.3530} & \textbf{0.2497} & \textbf{0.4562} \\
ICL+NH+CE+Judge Criteria &  0.3436 & 0.2383 & 0.4489 \\
\textbf{Merge2\&ICL+NH+CE+JC (Merge3)} &  \textbf{0.3553} & \textbf{0.2504} & \textbf{0.4602} \\

\hline
\end{tabular}
\begin{tablenotes}
\footnotesize
\item[1] \& means a child model merged from  parent models
\end{tablenotes}
\end{threeparttable}
\label{tab:merge}
\end{table}

\section{Conclusion}

This study presents a novel three-stage framework for fine-grained Chinese hate speech detection, integrating prompt engineering, supervised fine-tuning, and LLM merging. Through systematic experimentation on the STATE-ToxiCN benchmark, we demonstrate that our prompt-driven approach significantly enhances LLMs' capability to decode implicit hate patterns through structured semantic decomposition. The LLM merging algorithm effectively synthesizes complementary detection capabilities from different fine-tuned models. The final merged model exhibits robust performance in handling complex scenarios while maintaining generalization capabilities. The results highlight the potential of the merge-based approach in addressing language-specific challenges, contributing to safer and more inclusive online discourse environments.

\section*{Acknowledgements}

This research was supported by the National Natural Science Foundation of China (NSFC) under Grant 72071029 and 72231010.

\bibliographystyle{ccl}
\bibliography{ref}

\end{document}